\documentclass[runningheads]{llncs}
\usepackage{booktabs}
\usepackage[T1]{fontenc}
\usepackage{amsmath,amssymb}
\usepackage{hyperref}   
\usepackage{graphicx,verbatim}
\usepackage{diagbox}
\usepackage{xcolor}
\newcommand{\best}[1]{\textcolor{red}{#1}}  % SOTA
\newcommand{\runnerup}[1]{\underline{#1}}  % 第二名（黑白也清楚）
\newcommand{\sig}{\textsuperscript{*}}     % 显著性
\begin{document}
%
%\title{Brain-DiT: A Universal Multi-state fMRI Foundation Model via Diffusion Transformers}
\title{Brain-DiT: A Universal Multi-state fMRI Foundation Model with Metadata-Conditioned Pretraining}
\titlerunning{Brain-DiT}
% \begin{comment}  %% Removed for anonymized MICCAI submission

% \end{comment}

\author{Junfeng Xia\inst{1} \and Wenhao Ye\inst{1,2} \and Xuanye Pan\inst{1} \and Xinke Shen\inst{1} \and Mo Wang\inst{1}\sig \and Quanying Liu\inst{1}\sig}
\authorrunning{J. Xia et al.}
\institute{Department of Biomedical Engineering, Southern University of Science and Technology, China \and School of Biomedical Engineering, Shenzhen University, China \\
    \email{12250099@mail.sustech.edu.cn;liuqy@sustech.edu.cn}\\
    \textsuperscript{*}Co-corresponding authors}
\maketitle              % typeset the header of the contribution
\begin{abstract}
Current fMRI foundation models primarily rely on a limited range of brain states and mismatched pretraining tasks, restricting their ability to learn generalized representations across diverse brain states. We present \textit{Brain-DiT}, a universal multi-state fMRI foundation model pretrained on 349,898 sessions from 24 datasets spanning resting, task, naturalistic, disease, and sleep states. Unlike prior fMRI foundation models that rely on masked reconstruction in the raw-signal space or a latent space, \textit{Brain-DiT} adopts metadata-conditioned diffusion pretraining with a Diffusion Transformer (DiT), enabling the model to learn multi-scale representations that capture both fine-grained functional structure and global semantics. Across extensive evaluations and ablations on 7 downstream tasks, we find consistent evidence that diffusion-based generative pretraining is a stronger proxy than reconstruction or alignment, with metadata-conditioned pretraining further improving downstream performance by disentangling intrinsic neural dynamics from population-level variability. We also observe that downstream tasks exhibit distinct preferences for representational scale: ADNI classification benefits more from global semantic representations, whereas age/sex prediction comparatively relies more on fine-grained local structure. Code and parameters of Brain-DiT are available at \href{https://github.com/REDMAO4869/Brain-DiT}{Link}.
\end{abstract}

% Authors must provide keywords and are not allowed to remove this Keyword section.

% \end{abstract}
%
%
%

\section{Introduction}

% The human brain is a dynamical system that continuously transits between states, ranging from spontaneous rest to complex cognitive tasks, to flexibly support adaptive behaviors~\cite{shine_dynamics_2016,}.
The human brain can be viewed as a structurally constrained dynamical system whose large-scale functional organization continuously transitions across diverse states, ranging from spontaneous rest to complex cognitive tasks, to flexibly support adaptive cognition and behavior~\cite{shine_dynamics_2016,qu2024genetic}.

Capturing universal representations across this state spectrum is essential for decoding brain function, yet traditional approaches often fail to generalize across different brain states. Inspired by the transformative impact of self-supervised learning in computer vision and natural language processing, neuroimaging (such as fMRI, EEG and MEG data) is undergoing a paradigm shift: moving from task-specific models to large-scale foundation models pretrained on massive unlabeled datasets.

\begin{figure}[t]
\centering
\includegraphics[width=\textwidth]{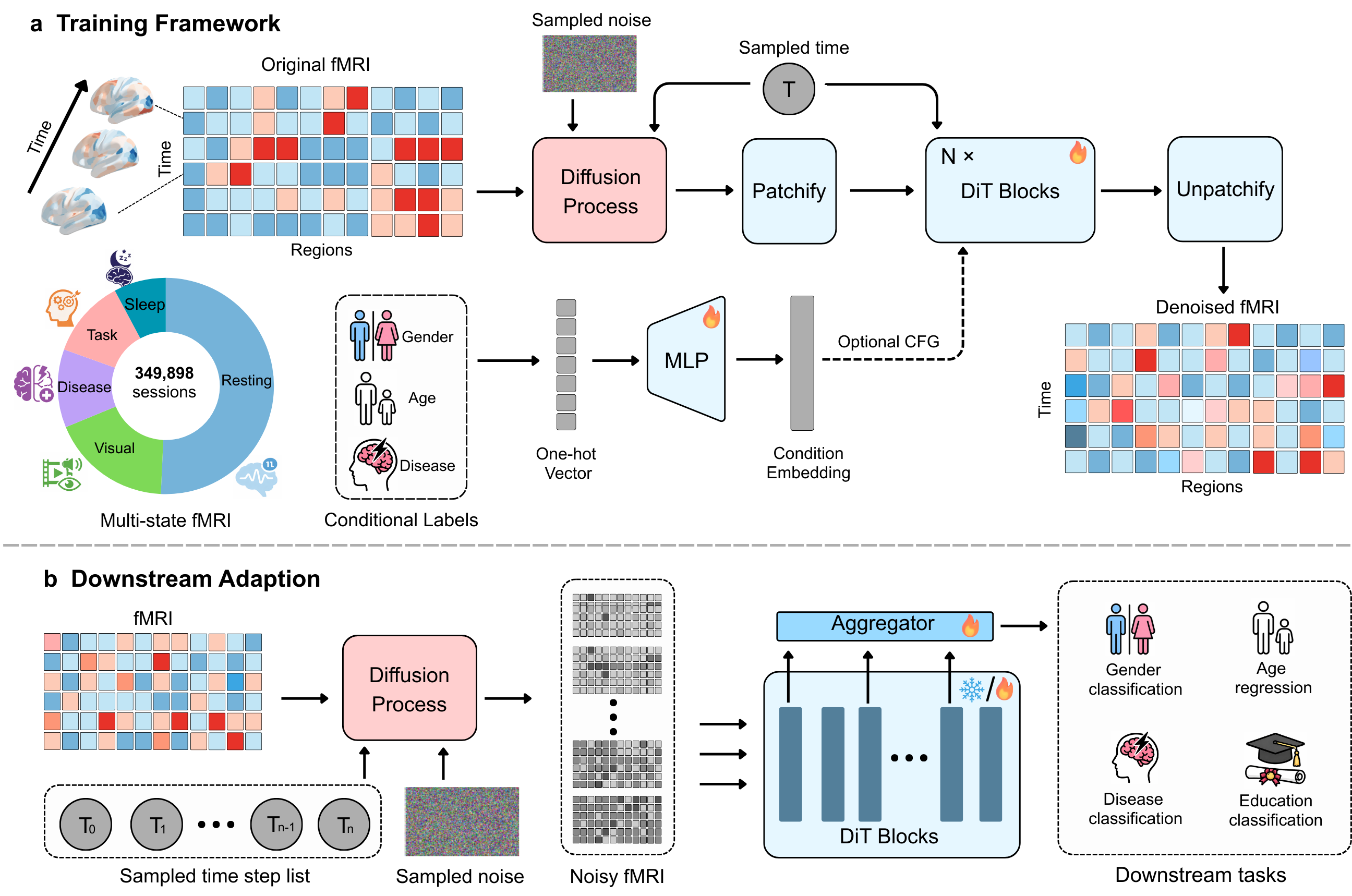}
\caption{\textbf{Brain-DiT framework: pretraining and downstream adaptation.}
(a) \textbf{Pretraining.} An ROI time-series window is noised at a random timestep $t$ and denoised by a conditional/unconditional DiT using $v$-prediction.
(b) \textbf{Downstream adaptation.} Multiple noisy views at selected timesteps are processed by the shared pretrained DiT. Token features from selected layers are aggregated to form a representation for downstream classification or regression.}
\label{fig1}
\end{figure}

% Recent efforts on fMRI foundation models have explored diverse proxy tasks, including masked autoencoding (MAE)~\cite{qu_uncovering_2024,caro_brainlm_2023}, joint-embedding predictive architectures (JEPA)~\cite{dong_brain-jepa_nodate} and contrastive learning~\cite{yang_brainmass_2024} to extract universal representations from neural signals at scale.

Recent efforts on fMRI foundation models have explored diverse proxy tasks, including masked autoencoding (MAE)~\cite{qu_uncovering_2024,caro_brainlm_2023,wang2026omni}, joint-embedding predictive architectures (JEPA)~\cite{dong_brain-jepa_nodate}, contrastive learning~\cite{yang_brainmass_2024}, and diffusion-based generative modeling of fMRI dynamics~\cite{xia2026brainworld}. However, existing fMRI foundation models exhibit limited robustness across diverse brain states, primarily due to three fundamental limitations. First, the pretraining distribution is restricted: most models are trained exclusively on resting-state data, limiting their exposure to the full spectrum of brain dynamics~\cite{caro_brainlm_2023,yang_brainmass_2024,dong_brain-jepa_nodate,wang_slim-brain_2026}.
% Recent work has shown that such narrow pretraining restricts models to state-specific statistics~\cite{wei_brainmoe_nodate}, compromising generalization to task-based or clinical populations. 
Second, the proxy tasks employed are often mismatched to learn universal representations: raw reconstruction may overfit to noise in low-SNR fMRI data~\cite{caro_brainlm_2023,qu_uncovering_2024}, while latent-space reconstruction requires careful architectural design to avoid representational collapse~\cite{wang_slim-brain_2026,dong_brain-jepa_nodate}. Consequently, substantial post-training adaptation is often required to achieve acceptable downstream performance. A third, largely unexplored limitation is the underutilization of clinical and demographic metadata during pretraining, which hinders the disentanglement of intrinsic neural dynamics from population-level variability.

\begin{figure}[t]
\centering
\includegraphics[width=\textwidth]{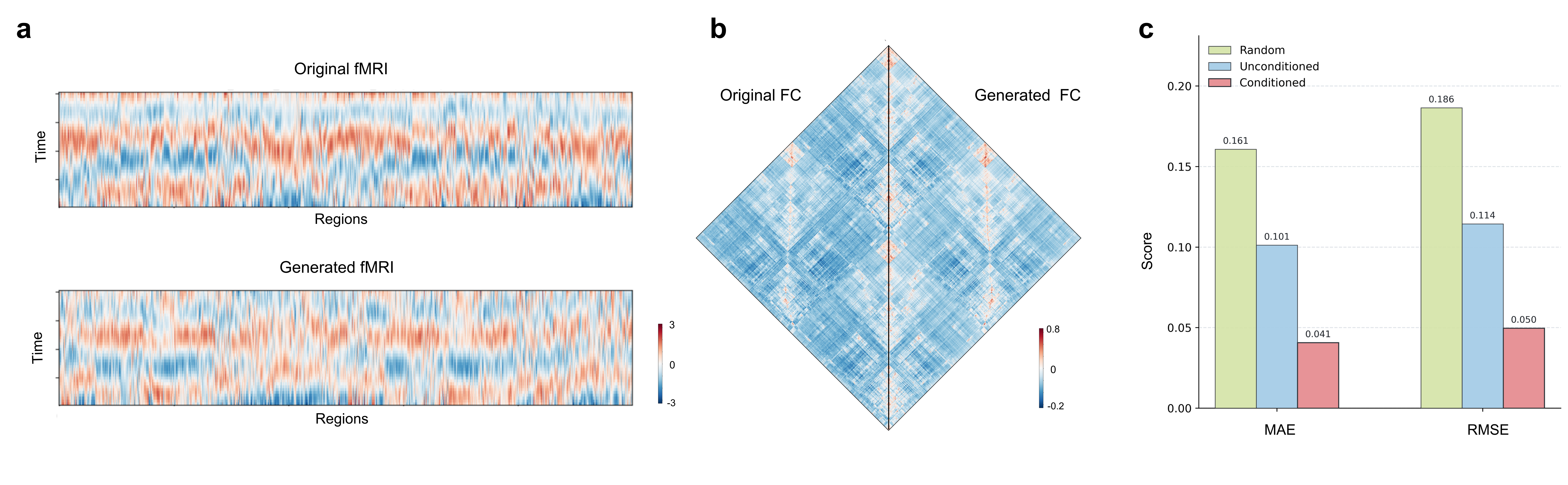}
\caption{\textbf{Metadata-conditioned generation of Brain-DiT.}
(a) \textbf{Original vs.\ generated fMRI signals.} Generated fMRI signals exhibit similar spatiotemporal patterns to the original fMRI.
(b) \textbf{Group-level FC comparison.} Empirical and synthetic group-level FC matrices for 200 metadata-conditioned virtual ASD subjects exhibit similar large-scale connectivity organization.
(c) \textbf{FC error metrics.} MAE and RMSE (lower is better) across three settings: random signal, unconditional generation, and metadata-conditioned generation. All results are computed on the ABIDE dataset.
}
\label{fig2}
\end{figure}

\begin{table}[t]
\caption{\textbf{Downstream performance on in-distribution (ID) and out-of-distribution (OOD) tasks.} Top: ID; bottom: OOD. Brain-DiT$_{\mathrm{AAL424}}$ and Brain-DiT$_{\mathrm{uncond}}$ are ablations. Best and runner-up results are highlighted in \best{red} and by underlining, respectively. * indicates $p<0.05$ vs.\ the best baseline (BrainLM/Brain-JEPA/BrainMass); no tests are performed among our variants.}
\label{table1}
\centering
\setlength{\tabcolsep}{1.0pt}
\renewcommand{\arraystretch}{1.15}
\footnotesize

\begin{tabular}{lccccc}
\toprule
\diagbox[width=6.5em]{\textbf{Model}}{\textbf{Dataset}} &
\textbf{HCP} &
\multicolumn{2}{c}{\textbf{ABIDE-AGE}} &
\multicolumn{2}{c}{\textbf{SALD}} \\
\cmidrule(lr){2-2}\cmidrule(lr){3-4}\cmidrule(lr){5-6}
& ACC\%$\uparrow$ & MSE$\downarrow$ & R$\uparrow$ & MSE$\downarrow$ & R$\uparrow$ \\
\midrule
BrainLM      & $62.71 \pm 4.43$ & $0.91 \pm 0.01$ & $0.24 \pm 0.02$ & $0.68 \pm 0.06$ & $0.62 \pm 0.06$ \\
Brain-JEPA    & $69.97 \pm 2.73$ & $0.98 \pm 0.07$ & $0.17 \pm 0.02$ & $1.13 \pm 0.11$ & $0.30 \pm 0.06$ \\
BrainMass    & $67.65 \pm 1.02$ & \best{$0.70 \pm 0.04$} & $0.50 \pm 0.04$ & $0.70 \pm 0.08$ & $0.63 \pm 0.06$ \\
\midrule
Brain-DiT$_{\mathrm{AAL424}}$  & \runnerup{$81.71 \pm 0.77$}  &  $0.80 \pm 0.02$ & $0.59 \pm 0.01$ & $0.53 \pm 0.02$ & $0.65 \pm 0.01$ \\
Brain-DiT$_{\mathrm{uncond}}$ & $80.98 \pm 1.22$ & $0.77 \pm 0.02$ & \runnerup{$0.60 \pm 0.02$} & \runnerup{$0.39 \pm 0.02$} & \runnerup{$0.75 \pm 0.01$} \\
\textbf{Brain-DiT} &
% \textbf{82.71} $\pm$ \textbf{0.79} & \textbf{0.75} $\pm$ \textbf{0.01}& \textbf{0.61} $\pm$ \textbf{0.01}& \textbf{0.33} $\pm$ \textbf{0.01}& \textbf{0.79} $\pm$ \textbf{0.01} \\
\best{82.71 $\pm$ 0.79\sig} & \runnerup{0.75 $\pm$ 0.01} & \best{0.61 $\pm$ 0.01\sig} & \best{0.33 $\pm$ 0.01\sig} & \best{0.79 $\pm$ 0.01\sig} \\
\bottomrule
\end{tabular}

\medskip

\begin{tabular}{lccccc}
\toprule
\diagbox[width=6.5em]{\textbf{Model}}{\textbf{Dataset}} &
\textbf{PPMI} &
\textbf{ADHD} &
\textbf{NKI-EDU} &
\multicolumn{2}{c}{\textbf{NKI-AGE}} \\
\cmidrule(lr){2-2}\cmidrule(lr){3-3}\cmidrule(lr){4-4}\cmidrule(lr){5-6}
& ACC\%$\uparrow$ & ACC\%$\uparrow$ & ACC\%$\uparrow$ & MSE$\downarrow$ & R$\uparrow$ \\
\midrule
BrainLM      & $54.86 \pm 1.20$ & $58.37 \pm 1.04$ & $60.46 \pm 0.73$ & $0.50 \pm 0.02$ & $0.68 \pm 0.01$ \\
Brain-JEPA    & $47.91 \pm 3.61$ & $54.89 \pm 4.98$ & $49.21 \pm 2.15$ & $1.03 \pm 0.08$ & $0.33 \pm 0.03$ \\
BrainMass    & $58.68 \pm 4.21$ & $57.21 \pm 3.45$ & $62.36 \pm 1.24$ & $0.60 \pm 0.06$ & $0.62 \pm 0.04$ \\
\midrule
Brain-DiT$_{\mathrm{AAL424}}$  &  $58.99 \pm 1.46$ & \runnerup{$62.38 \pm 0.67$} & $60.18 \pm 1.81$ &  $0.44 \pm 0.04$ & $0.77 \pm 0.02$ \\
Brain-DiT$_{\mathrm{uncond}}$ & \runnerup{$60.99 \pm 2.14$} & $62.06 \pm 0.36$ & \runnerup{$69.78 \pm 3.29$} & \runnerup{$0.40 \pm 0.02$} & \runnerup{$0.80 \pm 0.01$} \\
\textbf{Brain-DiT} &
\best{61.15 $\pm$ 1.84\sig}  & \best{62.53 $\pm$ 1.56\sig} & \best{69.92 $\pm$ 0.02\sig} & \best{0.38 $\pm$ 0.01\sig}  & \best{0.81 $\pm$ 0.01\sig} \\
\bottomrule
\end{tabular}

\end{table}

Diffusion models, which learn to reverse a gradual noising process, offer a promising solution to these challenges. First, by modeling the full data distribution rather than optimizing discriminative objectives, diffusion models characterize the complete probability manifold of neural signals, enabling the learning of richer, more generalizable features~\cite{mukhopadhyay_text-free_2024}. Second, the iterative denoising process provides a natural hierarchy of representations, with early timesteps capturing fine-grained local structure and later timesteps encoding abstract global semantics. Most importantly, diffusion models natively support conditional generation, allowing seamless integration of biological and clinical metadata as conditioning signals---thereby reducing the ``contextual blindness'' of prior paradigms.
%Diffusion models, which simulate the data creation process rather than memorizing local patterns, offer a promising solution to these challenges~\cite{seo_scalable_2025}. By modeling the full probability distribution of fMRI signals, diffusion models can extract deeper feature manifolds and enable conditional pretraining, allowing the incorporation of demographic and clinical metadata. This reduces the contextual blindness seen in previous approaches and enables a more nuanced understanding of brain dynamics~\cite{mukhopadhyay_text-free_2024}.
% Moreover, the reverse denoising trajectory offers a multi-scale parsing process, with higher-noise steps capturing coarse global structure and lower-noise steps refining fine-grained local details.
% By leveraging metadata-conditioned generation and extracting multi-scale representations from the reverse denoising process, this approach provides a potential solution to overcome representational fragility and capture universal brain dynamics.

Here, we present \textbf{Brain-DiT}, a universal multi-state foundation model that leverages these advantages through a metadata-conditioned Diffusion Transformer. Our main contributions are:
\begin{itemize}
\item \textbf{Multi-state generative pretraining paradigm.} We introduce the first fMRI foundation model trained jointly on multiple brain states, unifying 349,898 sessions from 24 datasets spanning resting, task, naturalistic, disease, and sleep conditions under a single generative objective.
\item \textbf{Metadata-conditioned pretraining.} We are the first to integrate demographic and clinical metadata as explicit conditioning signals during pretraining, enabling the model to disentangle intrinsic neural dynamics from population variability.
\item \textbf{Multi-scale feature aggregation.} We treat the denoising trajectory as a hierarchical feature extractor and aggregate representations across timesteps and layers, yielding embeddings that capture both fine-grained functional structure and abstract semantic information.
\end{itemize}

 %% removed for anonymized MICCAI submission.
    
    % The following acknowledgement and disclaimer sections can be removed for the double-blind review process.  If and when your paper is accepted, reinsert the acknowledgement and the disclaimer clause in your final camera-ready version.
    % IF you opted to include the acknowledgement and disclaimer sections, they will count towards the 8-page limit.

\begin{credits}
\end{credits}

\section{Methods}
\label{sec:methods}

\subsection{Problem Formulation and Overview}
% We study foundation model pretraining on multi-state ROI-based fMRI time series.
An fMRI sample is denoted by $\mathbf{x}\in\mathbb{R}^{T\times N}$, where $T$ is the number of time points and $N$ is the number of ROIs.
For patch-based Transformer modeling, we reshape $\mathbf{x}$ into an ROI$\times$time tensor $\mathbf{x}_0\in\mathbb{R}^{1\times N\times T}$ (channel $\times$ ROI $\times$ time).
Our pretraining objective is to learn a diffusion-based generative model for $p(\mathbf{x})$ by training a DiT denoiser, optionally conditioned on subject-level metadata.
% Multi-state fMRI exhibits distribution shifts across states and datasets, challenging state-robust representation learning from raw signals.We address this with diffusion pretraining, metadata conditioning, and trajectory aggregation via Layer Attention.

\subsection{Generative Diffusion Pretraining}
\label{subsec:pretraining}

\subsubsection{Diffusion and Training Process}
As shown in Fig.~\ref{fig1}\,(a), we adopt a DDPM forward process with a cosine noise schedule over $S$ diffusion steps ($S{=}1000$). For timestep $t\in\{1,\dots,S\}$, we sample Gaussian noise $\boldsymbol{\epsilon}\sim\mathcal{N}(\mathbf{0},\mathbf{I})$ and obtain:
\begin{equation}
\mathbf{x}_t \;=\; \alpha_t \mathbf{x}_0 \;+\; \sigma_t \boldsymbol{\epsilon},
\end{equation}
where $\alpha_t=\sqrt{\bar{\alpha}_t}$ and $\sigma_t=\sqrt{1-\bar{\alpha}_t}$ are precomputed schedule coefficients. 

Given the noised input $\mathbf{x}_t\in\mathbb{R}^{1\times N\times T}$ and timestep $t$, we learn a DiT directly in raw signal space. DiT first patchifies $\mathbf{x}_t$ using a $2$D convolution with patch size $p$ and stride $p$, producing a token sequence of length
\begin{equation}
L \;=\; (N/p)\cdot(T/p),
\end{equation}
with embedding dimension $D$. Tokens are processed by $K$ DiT blocks, and a linear head followed by an invertible unpatchify operation maps tokens back to a dense prediction in $\mathbb{R}^{1\times N\times T}$. We then train the model with $v$-prediction, where
\begin{equation}
\mathbf{v} \;=\; \alpha_t \boldsymbol{\epsilon} \;-\; \sigma_t \mathbf{x}_0.
\end{equation}
The DiT is trained to predict $\hat{\mathbf{v}}_\theta(\mathbf{x}_t,t,\mathbf{c})$ by minimizing:
\begin{equation}
\mathcal{L}_{\mathrm{diff}} \;=\;
\mathbb{E}_{\mathbf{x}_0,t,\boldsymbol{\epsilon}}
\left[
\left\|\hat{\mathbf{v}}_\theta(\mathbf{x}_t,t,\mathbf{c}) - \mathbf{v}\right\|_2^2
\right].
\end{equation}

\subsubsection{Metadata-Conditioned Pretraining}
\label{subsec:cond_injection}
Brain-DiT uses optional explicit metadata conditioning during pretraining.
Given subject metadata $\mathbf{c}$ (e.g., age, sex, diagnosis), we form a tabular vector with continuous values plus observation masks and one-hot diagnosis indicators, then map it to
\begin{equation}
\mathbf{y}_c = f_{\phi}(\mathbf{c}) \in \mathbb{R}^{d_c}.
\end{equation}
With timestep embedding $\mathbf{y}_t \in \mathbb{R}^{d_c}$, the condition is
\begin{equation}
\mathbf{y} = \mathbf{y}_t + \mathbf{y}_c,
\end{equation}
which is injected into every DiT block via AdaLN-Zero to condition the denoising process.
During training, CFG-style regularization replaces $\mathbf{y}_c$ with an unconditional embedding with probability $p_{\mathrm{drop}}$; samples with missing metadata are either dropped or treated as unconditional according to the missing-label policy.

% To regularize conditioning, we apply classifier-free condition dropout and replace $\mathbf{y}_c$ with an unconditional embedding with probability $p_{\mathrm{drop}}$. For samples with missing metadata, we keep them in training and use the unconditional embedding for those samples, while applying metadata conditioning to samples with available metadata.

\begin{table}[t]
\caption{\textbf{Downstream performance with a frozen backbone.} Best results are shown in \best{red}; underlined values denote the runner-up. * indicates $p<0.05$ vs.\ the best baseline.}
\label{table2}
\centering
\setlength{\tabcolsep}{1.6pt}
\renewcommand{\arraystretch}{1.15}
\footnotesize

\begin{tabular}{lccccc}
\toprule
\diagbox[width=6.5em]{\textbf{Model}}{\textbf{Dataset}} &
\textbf{ADHD} &
\textbf{SALD} &
\textbf{NKI-EDU} & 
\textbf{NKI-AGE} & \\
\cmidrule(lr){2-5}
& ACC\%$\uparrow$ & MSE$\downarrow$ & ACC\%$\uparrow$ & MSE$\downarrow$ \\
\midrule
BrainLM      & $53.57 \pm 2.77$& $0.80 \pm 0.01$ & $47.95 \pm 1.12$ & $0.76 \pm 0.02$ \\
Brain-JEPA    & $52.24 \pm 2.66$& $0.76 \pm 0.01$ & $38.69 \pm 0.23$ & $0.95 \pm 0.02$  \\
BrainMass    &  $54.89 \pm 3.38$ & $0.78 \pm 0.03$ & $54.89 \pm 3.38$ & $0.72 \pm 0.01$ \\
SlimBrain$_{\mathrm{jepa}}$   & \runnerup{$55.74 \pm 6.51$} & $0.72 \pm 0.01$ & $53.83 \pm 6.62$ & $0.89 \pm 0.02$ \\
SlimBrain$_{\mathrm{mae}}$    & $55.72 \pm 1.01$ & \runnerup{$0.71 \pm 0.07$} & \runnerup{$60.59 \pm 2.77$} & \runnerup{$0.68 \pm 0.01$} \\
\midrule
\textbf{Brain-DiT} &
\best{62.04 $\pm$ 1.22\sig} &
\best{0.67 $\pm$ 0.01\sig} &
\best{62.80 $\pm$ 0.63\sig} &
\best{0.58 $\pm$ 0.03\sig} & \\
\bottomrule
\end{tabular}
\end{table}

\subsection{Multi-scale Feature Aggregation}
\label{subsec:traj_agg}

% \subsubsection{Why Trajectory Features}
% Higher-noise timesteps emphasize global structure, while lower-noise timesteps preserve fine-grained temporal dynamics. Thus, multi-timestep representations can capture complementary information that is difficult to obtain from a single deterministic encoder.
\subsubsection{Feature Extraction}
Given a window $\mathbf{x}$, we construct $\mathbf{x}_0\in\mathbb{R}^{1\times N\times T}$ and choose a timestep list $\mathcal{T}=\{t_1,\dots,t_M\}$ (e.g., $\{0,50,100,150\}$) and a set of captured Transformer blocks $\mathcal{L}$ (e.g., the last few blocks).
For each $t\in\mathcal{T}$, we generate $\mathbf{x}_t$ using the same forward process:
\begin{equation}
\mathbf{x}_t \;=\; \alpha_t\mathbf{x}_0 \;+\; \sigma_t\boldsymbol{\epsilon}.
\end{equation}
In Fig.~\ref{fig1}\,(b), we pass $\mathbf{x}_t$ through the pretrained DiT and extract features from selected blocks $\ell\in\mathcal{L}$.
For each $(t,\ell)$, we mean-pool
$\mathbf{H}^{(\ell)}_t\in\mathbb{R}^{B\times L\times D}$ to obtain
\begin{equation}
\mathbf{e}_{t,\ell}=\mathrm{MeanPool}\!\left(\mathbf{H}^{(\ell)}_t\right)\in\mathbb{R}^{B\times D}.
\end{equation}
We then concatenate features across timesteps and layers:
\begin{equation}
\mathbf{E}=\mathrm{Concat}_{t\in\mathcal{T},\,\ell\in\mathcal{L}}\mathbf{e}_{t,\ell}
\in\mathbb{R}^{B\times Q\times D},\quad Q=|\mathcal{T}||\mathcal{L}|.
\end{equation}
% We forward $\mathbf{x}_t$ through the pretrained DiT and capture intermediate token features from blocks $\ell\in\mathcal{L}$. Let $\mathbf{H}^{(\ell)}_t\in\mathbb{R}^{B\times L\times D}$ denote the token matrix at layer $\ell$ for timestep $t$. We compute a pooled vector:
% \begin{equation}
% \mathbf{e}_{t,\ell} \;=\; \mathrm{MeanPool}\big(\mathbf{H}^{(\ell)}_t\big) \in \mathbb{R}^{B\times D}.
% \end{equation}
% Stacking all $(t,\ell)$ features yields:
% \begin{equation}
% \mathbf{E} \;=\; [\mathbf{e}_{t_1,\ell_1},\dots,\mathbf{e}_{t_M,\ell_{|\mathcal{L}|}}] \in \mathbb{R}^{B\times Q\times D},\quad Q=|\mathcal{T}||\mathcal{L}|.
% \end{equation}

\subsubsection{Aggregation}
Given $\mathbf{E}\in\mathbb{R}^{B\times Q\times D}$ with $Q=|\mathcal{T}||\mathcal{L}|$, we fuse multi-timestep, multi-layer features using a query-based aggregator module.
A learnable query $\mathbf{q}\in\mathbb{R}^{1\times 1\times D}$ is broadcast across the batch and used for cross-attention:
\begin{equation}
\mathbf{z},\,\mathbf{w}
=\mathrm{Attn}(\mathbf{q},\mathbf{E},\mathbf{E}),
\quad
\mathbf{z}\in\mathbb{R}^{B\times D},\;
\mathbf{w}\in\mathbb{R}^{B\times Q}.
\end{equation}
The fused representation $\mathbf{z}$ is fed to the downstream prediction head, while $\mathbf{w}$ provides the relative importance of each $(t,\ell)$ pair.

\section{Experiments and Results}

\subsection{Experimental Setup}
\textbf{Datasets and Preprocessing.} We utilize 24 multi-state datasets covering resting, task, disease, naturalistic, and sleep states, partitioned into ID and OOD sets. The ID pretraining corpus comprises 22 datasets: 9 resting-state (HCP~\cite{van_essen_wu-minn_2013}, CHCP~\cite{ge_increasing_2023}, ABCD~\cite{casey2018adolescent}, PIOP1~\cite{snoek_amsterdam_2021}, PIOP2~\cite{snoek_amsterdam_2021}, ISYB~\cite{gao_chinese_2022}, BHRC~\cite{de_oliveira_longitudinal_2025}, SALD~\cite{Wei2017StructuralAF}, SLIM~\cite{liu_longitudinal_2017}), 2 disease (ABIDE~\cite{di2014autism}, ADNI~\cite{jack_alzheimers_2008}), and 10 specialized states including task-based (HCP Task~\cite{van_essen_wu-minn_2013}, CHCP Task~\cite{ge_increasing_2023}, N\&N~\cite{nakai_quantitative_2020}), naturalistic (StudyForrest~\cite{hanke_studyforrest_2016}, Emo Film~\cite{morgenroth_emo-film_2024}, NSD~\cite{allen_massive_2022}, Things~\cite{hebart_things-data_2023}, CineBrain~\cite{gao_cinebrain_2025}, HCP Movie~\cite{van_essen_wu-minn_2013}), and sleep fMRI~\cite{sleep}. To validate generalization, the NKI~\cite{telesford_open-access_2023}, ADHD-200~\cite{adhd2012adhd}, and PPMI~\cite{marek_parkinson_2011} datasets are strictly reserved for OOD evaluation. All fMRI volumes were resampled to $2\,\text{mm}$ isotropic resolution and cropped/padded to $96 \times 96 \times 96$, with time series interpolated to a uniform TR of $0.72\,\text{s}$ (cubic B-spline). After global z-score normalization, voxel signals were parcellated into ROI time series using the Schaefer-1000~\cite{schaefer2018local} and AAL424 atlases~\cite{nemati2020unique}.

\textbf{Baselines and Evaluation.} We compare Brain-DiT with ROI and voxel baselines (BrainLM, Brain-JEPA, BrainMass, SlimBrain$_{\mathrm{mae}}$, SlimBrain$_{\mathrm{jepa}}$), covering MAE, JEPA, and contrastive learning. Evaluation metrics include Accuracy for classification, and Mean Squared Error (MSE) and Pearson’s $r$ for regression. Generative fidelity is quantified by comparing synthesized and empirical functional connectivity (FC) patterns via MSE and RMSE.
% \textbf{Baselines and Evaluation.} We compare Brain-DiT with ROI-based foundation models (BrainLM, Brain-JEPA, BrainMass) and voxel-level paradigms (SlimBrain$_{\mathrm{mae}}$, SlimBrain$_{\mathrm{jepa}}$), covering MAE, JEPA, and contrastive learning. Evaluation metrics are Accuracy for classification, and Mean Squared Error (MSE) and Pearson’s $r$ for regression. Generative fidelity is quantified by MSE and RMSE between synthetic and empirical functional connectivity (FC) patterns.

\begin{figure}[t]
\centering
\includegraphics[width=\textwidth]{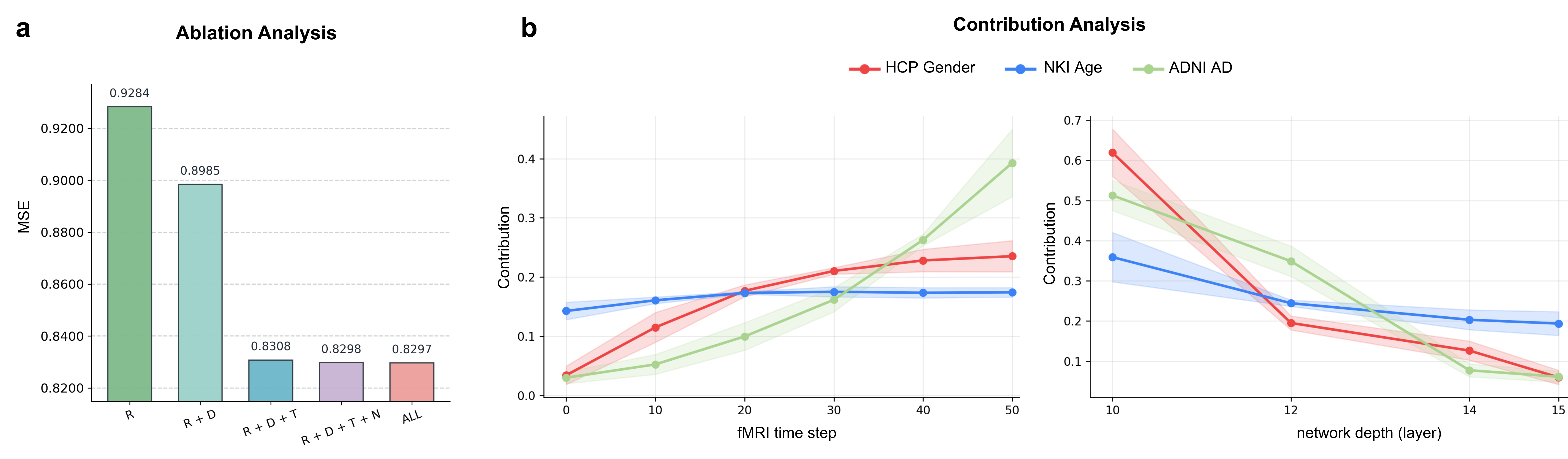}
\caption{\textbf{State diversity ablation and contribution analysis.}
(a) \textbf{State diversity ablation.} Mixtures of multi-state fMRI data are used during pretraining with a fixed pretraining budget ($\sim$30k sessions): R (resting), +D (disease), +T (task), +N (naturalistic), and ALL (the full multi-state corpus). %We show that downstream error decreases as state diversity increases. 
(b) \textbf{Contribution analysis.} Aggregator weights show contributions across timesteps (left) and network depth (right), reflecting global (later/noisier) vs.\ local (earlier/cleaner) features.}
\label{fig3}
\end{figure}

\textbf{Implementation Details.} Brain-DiT uses a DiT backbone ($d_{\text{model}}{=}1024$, 16 layers, 8 heads, patch size 4) and is trained on Schaefer-1000 ROI windows ($T{=}40$) with a cosine DDPM schedule (1,000 steps) and a $v$-prediction objective. We train for 150 epochs on 6 NVIDIA A800 (80GB) GPUs with AdamW (LR $1\mathrm{e}{-4}$, WD $1\mathrm{e}{-4}$; batch size 36/GPU, mixed precision) and inject age/sex/disease via AdaLN ($d_{\text{cond}}{=}256$) with optional 10\% condition dropout.

\subsection{Generative Fidelity}
% Before evaluating downstream task performance, we first examine whether Brain-DiT has learned meaningful representations via generative experiments.
Before evaluating downstream tasks, we first verify that Brain-DiT learns meaningful representations through generative experiments. As shown in Fig.~\ref{fig2}\,(a), we generated fMRI signals conditioned on specific subject metadata from the ABIDE dataset. The generated BOLD responses exhibit a high temporal correlation with the empirical ground truth, indicating that the model successfully generates individual-level neural dynamics guided by metadata. To further assess population-level fidelity, we generated fMRI signals for 200 virtual ASD subjects and computed group-level FC. The synthetic FC topography closely matches the empirical FC of the real cohort, and quantitative metrics (MSE/RMSE) show that conditional generation outperforms both unconditional and random-conditioning baselines. These results suggest that Brain-DiT learns subject-generalizable representations and supports controllable conditional generation.

\subsection{Downstream Task Performance}
As shown in Table~\ref{table1}, Brain-DiT outperforms most baselines on both ID and OOD benchmarks, with improvements that are significant against the best baseline. For fair comparison, we report two variants: Brain-DiT$_{\text{AAL424}}$, aligned with the AAL424 atlas used by BrainLM, and Brain-DiT$_{\text{uncond}}$, an unconditional ablation. 
% Even these variants remain strong; for example, \textbf{Brain-DiT$_{\text{AAL424}}$} reaches $81.71\%$ accuracy on HCP, compared with the best baseline Brain-JEPA ($69.97\%$). Moreover, the full model improves over \textbf{Brain-DiT$_{\text{uncond}}$} (HCP ACC: $80.98\%\!\rightarrow\!82.71\%$; SALD $r$: $0.75\!\rightarrow\!0.79$), highlighting the benefit of metadata conditioning for generalization.
Even the ablated variants remain strong. For example, Brain-DiT$_{\text{AAL424}}$ achieves $81.71\%$ accuracy on HCP, outperforming the best baseline (Brain-JEPA, $69.97\%$). Moreover, the full model outperforms Brain-DiT$_{\text{uncond}}$ across benchmarks, highlighting the benefit of metadata conditioning for generalization.

\subsection{Representation Quality}
We assess representation quality by freezing the pretrained backbone and training a lightweight linear head as shown in Table~\ref{table2}. Despite using ROI-level inputs, Brain-DiT surpasses \textbf{voxel-level} MAE/JEPA baselines significantly; for instance, it achieves $62.80\%$ on NKI-EDU versus $60.59\%$ for SlimBrain$_{\mathrm{mae}}$. This advantage persists under distribution shift, with Brain-DiT reaching $62.04\%$ accuracy on ADHD-200 and 0.67 MSE on SALD.

\subsection{Interpretability Analysis}
\textbf{State diversity drives downstream gains.}
To isolate the effect of state diversity, we fix the pretraining set size at 30k sessions and vary only the state mixture. As shown in Fig.~\ref{fig3}\,(a), downstream performance improves consistently as state coverage increases, suggesting that the gains mainly come from broader brain-state coverage rather than a larger sample size.

\noindent\textbf{Global--local contributions across timesteps and depth.}
% We analyze contributions across a set of discrete timesteps (e.g., $t=0$--$50$) and network depth, with larger $t$ indicating higher noise. From aggregator weights $w_{t,\ell}$, we compute timestep- and layer-level contribution ratios by summing over $\ell$ or $t$ and normalizing. 
We analyze step- and depth-wise contributions using normalized aggregator weights $w_{t,\ell}$, where larger $t$ indicates higher noise. Later timesteps emphasize coarse, network-level structure, whereas earlier steps preserve finer-grained variations. As shown in Fig.~\ref{fig3}\,(b), ADNI shows the strongest preference for later timesteps, consistent with system-level network disruptions in Alzheimer’s disease; age prediction is more evenly distributed, consistent with age effects spanning both global and local connectivity, with gender in between. Across depth, contributions concentrate in earlier-to-intermediate layers and decay in deeper blocks, with a sharper drop for disease classification, consistent with prior observations.
% ~\cite{lan2025diffusion}
% We analyze contributions across diffusion timesteps and network depth, where larger $t$ corresponds to higher noise, using normalized aggregator weights $w_{t,\ell}$. Later timesteps capture coarser network-level structure, whereas earlier timesteps retain finer-grained fluctuations. As shown in Fig.~\ref{fig3},(b), ADNI places the strongest weight on later timesteps, age prediction is more evenly distributed across timesteps, and gender lies in between, suggesting different preferences for global vs.\ local information. Across depth, contributions concentrate in early-to-mid layers and decline in deeper blocks, with a sharper drop for disease classification, in line with prior observations~\cite{lan2025diffusion}.

\section{Conclusion}
% We presented Brain-DiT, a universal multi-state fMRI foundation model with diffusion-based generative pretraining, optional metadata conditioning, and multi-scale feature aggregation.

We present Brain-DiT, a universal multi-state fMRI foundation model with metadata-conditioned pretraining to capture complex brain dynamics and multi-scale aggregation to improve downstream generalization. Brain-DiT outperforms representative baselines on ID and OOD benchmarks and remains strong under frozen-backbone transfer. Conditional generation reproduces realistic subject-level fMRI dynamics and preserves cohort-level FC patterns under metadata conditioning. Contribution analysis suggests that multi-state pretraining yields multi-scale representations, with diffusion timesteps and layers providing complementary signals for downstream tasks. Future work will extend Brain-DiT with richer metadata and broader state coverage, while improving scaling efficiency for longer sequences and higher-resolution inputs.

\bibliographystyle{splncs04}
\bibliography{bibs/dataset,bibs/story,bibs/techniche}

@article{gao_cinebrain_2025,
  title={CineBrain: A large-scale multi-modal brain dataset during naturalistic audiovisual narrative processing},
  author={Gao, Jianxiong and Liu, Yichang and Yang, Baofeng and Feng, Jianfeng and Fu, Yanwei},
  journal={arXiv preprint arXiv:2503.06940},
  year={2025}
}

@article{snoek_amsterdam_2021,
  title={The Amsterdam Open MRI Collection, a set of multimodal MRI datasets for individual difference analyses},
  author={Snoek, Lukas and van der Miesen, Maite M and Beemsterboer, Tinka and Van Der Leij, Andries and Eigenhuis, Annemarie and Steven Scholte, H},
  journal={Scientific data},
  volume={8},
  number={1},
  pages={85},
  year={2021},
  publisher={Nature Publishing Group UK London}
}

@article{gao_chinese_2022,
  title={A Chinese multi-modal neuroimaging data release for increasing diversity of human brain mapping},
  author={Gao, Peng and Dong, Hao-Ming and Liu, Si-Man and Fan, Xue-Ru and Jiang, Chao and Wang, Yin-Shan and Margulies, Daniel and Li, Hai-Fang and Zuo, Xi-Nian},
  journal={Scientific Data},
  volume={9},
  number={1},
  pages={286},
  year={2022},
  publisher={Nature Publishing Group UK London}
}

@article{ge_increasing_2023,
  title={Increasing diversity in connectomics with the Chinese Human Connectome Project},
  author={Ge, Jianqiao and Yang, Guoyuan and Han, Meizhen and Zhou, Sizhong and Men, Weiwei and Qin, Lang and Lyu, Bingjiang and Li, Hai and Wang, Haobo and Rao, Hengyi and others},
  journal={Nature Neuroscience},
  volume={26},
  number={1},
  pages={163--172},
  year={2023},
  publisher={Nature Publishing Group US New York}
}

@article{di2014autism,
  title={The autism brain imaging data exchange: towards a large-scale evaluation of the intrinsic brain architecture in autism},
  author={Di Martino, Adriana and Yan, Chao-Gan and Li, Qingyang and Denio, Erin and Castellanos, Francisco X and Alaerts, Kaat and Anderson, Jeffrey S and Assaf, Michal and Bookheimer, Susan Y and Dapretto, Mirella and others},
  journal={Molecular psychiatry},
  volume={19},
  number={6},
  pages={659--667},
  year={2014},
  publisher={Nature Publishing Group}
}

@article{van_essen_wu-minn_2013,
  title = {{The {WU}-{Minn} Human Connectome Project: An Overview}},
  author={Van Essen, David C and Smith, Stephen M and Barch, Deanna M and Behrens, Timothy EJ and Yacoub, Essa and Ugurbil, Kamil and Wu-Minn HCP Consortium and others},
  journal={Neuroimage},
  volume={80},
  pages={62--79},
  year={2013},
  publisher={Elsevier}
}

@article{marek_parkinson_2011,
  title={The Parkinson progression marker initiative (PPMI)},
  author={Marek, Kenneth and Jennings, Danna and Lasch, Shirley and Siderowf, Andrew and Tanner, Caroline and Simuni, Tanya and Coffey, Chris and Kieburtz, Karl and Flagg, Emily and Chowdhury, Sohini and others},
  journal={Progress in neurobiology},
  volume={95},
  number={4},
  pages={629--635},
  year={2011},
  publisher={Elsevier}
}

@article{jack_alzheimers_2008,
  title={The Alzheimer's disease neuroimaging initiative (ADNI): MRI methods},
  author={Jack Jr, Clifford R and Bernstein, Matt A and Fox, Nick C and Thompson, Paul and Alexander, Gene and Harvey, Danielle and Borowski, Bret and Britson, Paula J and L. Whitwell, Jennifer and Ward, Chadwick and others},
  journal={Journal of Magnetic Resonance Imaging: An Official Journal of the International Society for Magnetic Resonance in Medicine},
  volume={27},
  number={4},
  pages={685--691},
  year={2008},
  publisher={Wiley Online Library}
}

@article{de_oliveira_longitudinal_2025,
  title={Longitudinal patterns of disordered eating behaviors in children and adolescents from the Brazilian High-Risk Cohort study for mental conditions},
  author={de Oliveira, Iara Peixoto and Fernand{\'e}z, Ana C and Salum, Giovanni A and Gadelha, Ary and Pan, Pedro Mario and Miguel, Eur{\'\i}pedes Constantino and Mograbi, Daniel C and Bado, Patricia},
  journal={Brazilian Journal of Psychiatry},
  volume={47},
  pages={e20243867},
  year={2025},
  publisher={SciELO Brasil}
}

@article{telesford_open-access_2023,
  title={An open-access dataset of naturalistic viewing using simultaneous EEG-fMRI},
  author={Telesford, Qawi K and Gonzalez-Moreira, Eduardo and Xu, Ting and Tian, Yiwen and Colcombe, Stanley J and Cloud, Jessica and Russ, Brian E and Falchier, Arnaud and Nentwich, Maximilian and Madsen, Jens and others},
  journal={Scientific Data},
  volume={10},
  number={1},
  pages={554},
  year={2023},
  publisher={Nature Publishing Group UK London}
}

@ARTICLE{adhd2012adhd,
AUTHOR={Milham, Michael P. and Fair, Damien  and Mennes, Maarten  and Mostofsky, Stewart H.},
TITLE={The adhd-200 consortium: a model to advance the translational potential of neuroimaging in clinical neuroscience},
JOURNAL={Frontiers in Systems Neuroscience},
VOLUME={Volume 6 - 2012},
YEAR={2012},
ISSN={1662-5137},
}

@misc{sleep,
  author = {Yameng Gu AND Feng Han AND Lucas E. Sainburg AND Margeaux M. Schade AND Xiao Liu},
  title = {"Simultaneous EEG and fMRI signals during sleep from humans"},
  year = {2026},
  doi = {doi:10.18112/openneuro.ds003768.v1.0.13},
  publisher = {OpenNeuro}
}

@article{allen_massive_2022,
  title={A massive {7T} {fMRI} dataset to bridge cognitive neuroscience and artificial intelligence},
  author={Allen, Emily J and St-Yves, Ghislain and Wu, Yihan and Breedlove, Jesse L and Prince, Jacob S and Dowdle, Logan T and Nau, Matthias and Caron, Brad and Pestilli, Franco and Charest, Ian and others},
  journal={Nature neuroscience},
  volume={25},
  number={1},
  pages={116--126},
  year={2022},
  publisher={Nature Publishing Group US New York}
}

@article{hebart_things-data_2023,
  title={THINGS-data, a multimodal collection of large-scale datasets for investigating object representations in human brain and behavior},
  author={Hebart, Martin N and Contier, Oliver and Teichmann, Lina and Rockter, Adam H and Zheng, Charles Y and Kidder, Alexis and Corriveau, Anna and Vaziri-Pashkam, Maryam and Baker, Chris I},
  journal={Elife},
  volume={12},
  pages={e82580},
  year={2023},
  publisher={eLife Sciences Publications, Ltd}
}

@article{hanke_studyforrest_2016,
  title={A studyforrest extension, simultaneous fMRI and eye gaze recordings during prolonged natural stimulation},
  author={Hanke, Michael and Adelh{\"o}fer, Nico and Kottke, Daniel and Iacovella, Vittorio and Sengupta, Ayan and Kaule, Falko R and Nigbur, Roland and Waite, Alexander Q and Baumgartner, Florian and Stadler, J{\"o}rg},
  journal={Scientific data},
  volume={3},
  number={1},
  pages={160092},
  year={2016},
  publisher={Nature Publishing Group}
}

@article{nakai_quantitative_2020,
  title={Quantitative models reveal the organization of diverse cognitive functions in the brain},
  author={Nakai, Tomoya and Nishimoto, Shinji},
  journal={Nature communications},
  volume={11},
  number={1},
  pages={1142},
  year={2020},
  publisher={Nature Publishing Group UK London}
}

@article{liu_longitudinal_2017,
  title={Longitudinal test-retest neuroimaging data from healthy young adults in southwest China},
  author={Liu, Wei and Wei, Dongtao and Chen, Qunlin and Yang, Wenjing and Meng, Jie and Wu, Guorong and Bi, Taiyong and Zhang, Qinglin and Zuo, Xi-Nian and Qiu, Jiang},
  journal={Scientific data},
  volume={4},
  number={1},
  pages={170017},
  year={2017},
  publisher={Nature Publishing Group}
}

@article{Wei2017StructuralAF,
	title = {Structural and functional brain scans from the cross-sectional {Southwest} {University} adult lifespan dataset},
	volume = {5},
	issn = {2052-4463},
	number = {1},
	journal = {Scientific Data},
	author = {Wei, Dongtao and Zhuang, Kaixiang and Ai, Lei and Chen, Qunlin and Yang, Wenjing and Liu, Wei and Wang, Kangcheng and Sun, Jiangzhou and Qiu, Jiang},
	month = jul,
	year = {2018},
	pages = {180134},
}

@article{morgenroth_emo-film_2024,
  title={Emo-FilM: a multimodal dataset for affective neuroscience using naturalistic stimuli},
  author={Morgenroth, Elenor and Moia, Stefano and Vilaclara, Laura and Fournier, Raphael and Muszynski, Michal and Ploumitsakou, Maria and Almat{\'o}-Bellavista, Marina and Vuilleumier, Patrik and Van De Ville, Dimitri},
  journal={Scientific Data},
  volume={12},
  number={1},
  pages={684},
  year={2025},
  publisher={Nature Publishing Group UK London}
}

@article{casey2018adolescent,
  title={The adolescent brain cognitive development (ABCD) study: imaging acquisition across 21 sites},
  author={Casey, Betty Jo and Cannonier, Tariq and Conley, May I and Cohen, Alexandra O and Barch, Deanna M and Heitzeg, Mary M and Soules, Mary E and Teslovich, Theresa and Dellarco, Danielle V and Garavan, Hugh and others},
  journal={Developmental cognitive neuroscience},
  volume={32},
  pages={43--54},
  year={2018},
  publisher={Elsevier}
}

@article{schaefer2018local,
  title={Local-global parcellation of the human cerebral cortex from intrinsic functional connectivity MRI},
  author={Schaefer, Alexander and Kong, Ru and Gordon, Evan M and Laumann, Timothy O and Zuo, Xi-Nian and Holmes, Avram J and Eickhoff, Simon B and Yeo, BT Thomas},
  journal={Cerebral cortex},
  volume={28},
  number={9},
  pages={3095--3114},
  year={2018},
  publisher={Oxford University Press}
}

@article{nemati2020unique,
  title={A unique brain connectome fingerprint predates and predicts response to antidepressants},
  author={Nemati, Samaneh and Akiki, Teddy J and Roscoe, Jeremy and Ju, Yumeng and Averill, Christopher L and Fouda, Samar and Dutta, Arpan and McKie, Shane and Krystal, John H and Deakin, JF William and others},
  journal={IScience},
  volume={23},
  number={1},
  year={2020},
  publisher={Elsevier}
}

@article{shine_dynamics_2016,
  title={The dynamics of functional brain networks: integrated network states during cognitive task performance},
  author={Shine, James M and Bissett, Patrick G and Bell, Peter T and Koyejo, Oluwasanmi and Balsters, Joshua H and Gorgolewski, Krzysztof J and Moodie, Craig A and Poldrack, Russell A},
  journal={Neuron},
  volume={92},
  number={2},
  pages={544--554},
  year={2016},
  publisher={Elsevier}
}

@inproceedings{qu_uncovering_2024,
  title={Uncovering cognitive taskonomy through transfer learning in masked autoencoder-based fMRI reconstruction},
  author={Qu, Youzhi and Xia, Junfeng and Jian, Xinyao and Li, Wendu and Peng, Kaining and Liang, Zhichao and Wu, Haiyan and Liu, Quanying},
  booktitle={International Workshop on Human Brain and Artificial Intelligence},
  pages={35--50},
  year={2024},
  organization={Springer}
}

@inproceedings{qu2024genetic,
  title={A Genetic Algorithms for Optimizing Structural Brain Network Across Cognitive Tasks},
  author={Qu, Youzhi and Lit, Wendu and Xia, Junfeng and Tang, Jiahao and Peng, Kaining and Liang, Zhichao and Wu, Haiyan and Liu, Quanying},
  booktitle={2024 China Automation Congress (CAC)},
  pages={5210--5215},
  year={2024},
  organization={IEEE}
}

@article{wang2026omni,
  title={Omni-fMRI: A Universal Atlas-Free fMRI Foundation Model},
  author={Wang, Mo and Ye, Wenhao and Xia, Junfeng and Zhang, Junxiang and Pan, Xuanye and Xu, Minghao and Deng, Haotian and Wen, Hongkai and Liu, Quanying},
  journal={arXiv preprint arXiv:2601.23090},
  year={2026}
}

@article{xia2026brainworld,
  title={BrainWorld: A Structural-Prior-Conditioned Generative Model for Whole-Brain 4D fMRI Dynamics},
  author={Xia, Junfeng and Ye, Wenhao and Zhang, Junxiang and Pan, Xuanye and Wang, Mo and Liu, Quanying},
  journal={arXiv preprint arXiv:2606.17742},
  year={2026}
}

@inproceedings{caro_brainlm_2023,
  title={BrainLM: A foundation model for brain activity recordings},
  author={Ortega Caro, Josue and de Oliveira Fonseca, Antonio Henrique and Rizvi, Syed and Rosati, Matteo and Averill, Christopher and Cross, James and Mittal, Prateek and Zappala, Emanuele and Dhodapkar, Rahul and Abdallah, Chadi and others},
  booktitle={International Conference on Learning Representations},
  volume={2024},
  pages={565--576},
  year={2024}
}

@article{dong_brain-jepa_nodate,
  title={Brain-jepa: Brain dynamics foundation model with gradient positioning and spatiotemporal masking},
  author={Dong, Zijian and Li, Ruilin and Wu, Yilei and Nguyen, Thuan Tinh and Chong, Joanna and Ji, Fang and Tong, Nathanael and Chen, Christopher and Zhou, Juan Helen},
  journal={Advances in Neural Information Processing Systems},
  volume={37},
  pages={86048--86073},
  year={2024}
}

@article{yang_brainmass_2024,
  title={Brainmass: Advancing brain network analysis for diagnosis with large-scale self-supervised learning},
  author={Yang, Yanwu and Ye, Chenfei and Su, Guinan and Zhang, Ziyao and Chang, Zhikai and Chen, Hairui and Chan, Piu and Yu, Yue and Ma, Ting},
  journal={IEEE transactions on medical imaging},
  volume={43},
  number={11},
  pages={4004--4016},
  year={2024},
  publisher={IEEE}
}

@inproceedings{mukhopadhyay_text-free_2024,
  title={Do text-free diffusion models learn discriminative visual representations?},
  author={Mukhopadhyay, Soumik and Gwilliam, Matthew and Yamaguchi, Yosuke and Agarwal, Vatsal and Padmanabhan, Namitha and Swaminathan, Archana and Zhou, Tianyi and Ohya, Jun and Shrivastava, Abhinav},
  booktitle={European Conference on Computer Vision},
  pages={253--272},
  year={2024},
  organization={Springer}
}

@article{wang_slim-brain_2026,
  title={{SLIM-Brain}: A {Data-and} {Training-Efficient} {Foundation} {Model} for {fMRI} {Data} {Analysis}},
  author={Wang, Mo and Xia, Junfeng and Ye, Wenhao and Liu, Enyu and Peng, Kaining and Feng, Jianfeng and Liu, Quanying and Wen, Hongkai},
  journal={arXiv preprint arXiv:2512.21881},
  year={2025}
}
\end{document}